\documentclass[11pt]{article}

\usepackage[final]{acl}

\usepackage{xcolor}
\usepackage{listings}
\usepackage{amssymb}
\definecolor{codebg}{RGB}{248,249,250}
\definecolor{codekw}{RGB}{0,92,197}
\definecolor{codestr}{RGB}{34,134,58}
\definecolor{codecom}{RGB}{135,135,135}
\definecolor{codeapi}{RGB}{136,57,178}
\newcommand{\cmark}{\textcolor{green!55!black}{\checkmark}}
\newcommand{\pmark}{\textcolor{black!45}{$\circ$}}
\newcommand{\xmark}{\textcolor{red!65!black}{$\times$}}

\definecolor{codebg2}{RGB}{238,239,242}

\lstdefinestyle{appendixcode}{
    basicstyle=\ttfamily\fontsize{5.1pt}{5.45pt}\selectfont,
    numbers=none,
    escapeinside={(*@}{@*)},
    frame=single,
    framerule=0.25pt,
    rulecolor=\color{black!35},
    backgroundcolor=\color{codebg2},
    xleftmargin=14pt,
    xrightmargin=1pt,
    framexleftmargin=1pt,
    framexrightmargin=1pt,
    aboveskip=1pt,
    belowskip=1pt,
    breaklines=true,
    breakatwhitespace=false,
    breakindent=4pt,
    columns=fullflexible,
    keepspaces=true,
    showstringspaces=false
}

\newcounter{clnum}

\usepackage{times}
\usepackage{latexsym}
\usepackage[T1]{fontenc}
\usepackage[utf8]{inputenc}
\usepackage{microtype}
\usepackage{inconsolata}
\usepackage{graphicx}
\usepackage{array}

\title{\raisebox{-0.20\height}{\includegraphics[height=1.4em]{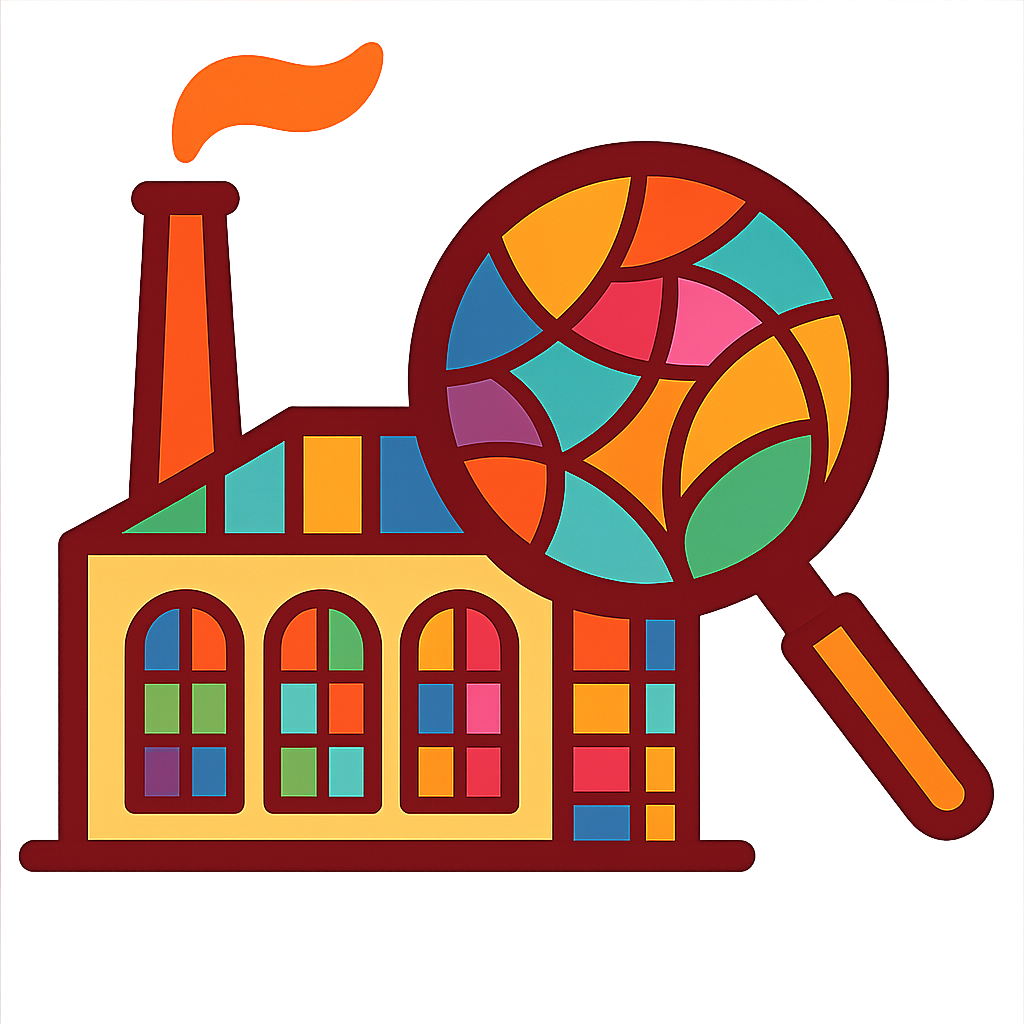}}\,\texttt{MURANO}: Design, Run, and Reproduce Mechanistic Interpretability Experiments as Composable Pipelines}

\author{
  \textbf{Alireza Bayat Makou\textsuperscript{1,*}},
  \textbf{Emirhan Böge\textsuperscript{1,4,*}},
  \textbf{Phu Gia Hoang\textsuperscript{1,*}},
  \textbf{Federico Tiblias\textsuperscript{1,2,3,*}}
\\
  \textbf{Jingcheng Niu\textsuperscript{1}},
  \textbf{Subhabrata Dutta\textsuperscript{1,2}},
  \textbf{Richard Eckart de Castilho\textsuperscript{1}},
  \textbf{Iryna Gurevych\textsuperscript{1,2,4}}
\\
   \small{\textbf{Correspondence:} \href{mailto:alireza.makou@tu-darmstadt.de}{alireza.makou@tu-darmstadt.de}
 }
\\
  {\small\textsuperscript{1}Ubiquitous Knowledge Processing (UKP) Lab, Technical University of Darmstadt, Germany}\\
  {\small\textsuperscript{2}National Research Center for Applied Cybersecurity ATHENE, Germany}\\
  {\small\textsuperscript{3}Zuse School ELIZA, Technical University of Darmstadt}\\
  {\small\textsuperscript{4}Cluster of Excellence ``Reasonable Artificial Intelligence'' (RAI), hessian.AI, Germany}
}

\begin{document}
\maketitle
\begin{abstract}
This paper presents Murano, an open source framework for designing, running, and reproducing mechanistic interpretability studies of large language models, intended for researchers across disciplines. These studies often combine loading, recording, attribution, intervention, and evaluation, while existing libraries tend to focus on different parts of this workflow. As a result, researchers using several libraries may need to adapt outputs from one for use by another. To bridge this gap, Murano represents operations from these five areas as composable steps. Steps exchange named result artifacts and declare the inputs they require and the outputs they produce. A pipeline executes its steps in the order supplied, and Murano uses canonical addresses when component identities pass between operations. Murano builds on existing interpretability and machine learning libraries. We demonstrate Murano through two reproductions of established interpretability studies and one illustrative sparse autoencoder case study.\footnote{\url{https://github.com/UKPLab/murano}\\\textsuperscript{*}Equal contribution.}
\end{abstract}

\section{Introduction}

\begin{figure*}[tp]
  \centering
  \includegraphics[width=\textwidth]{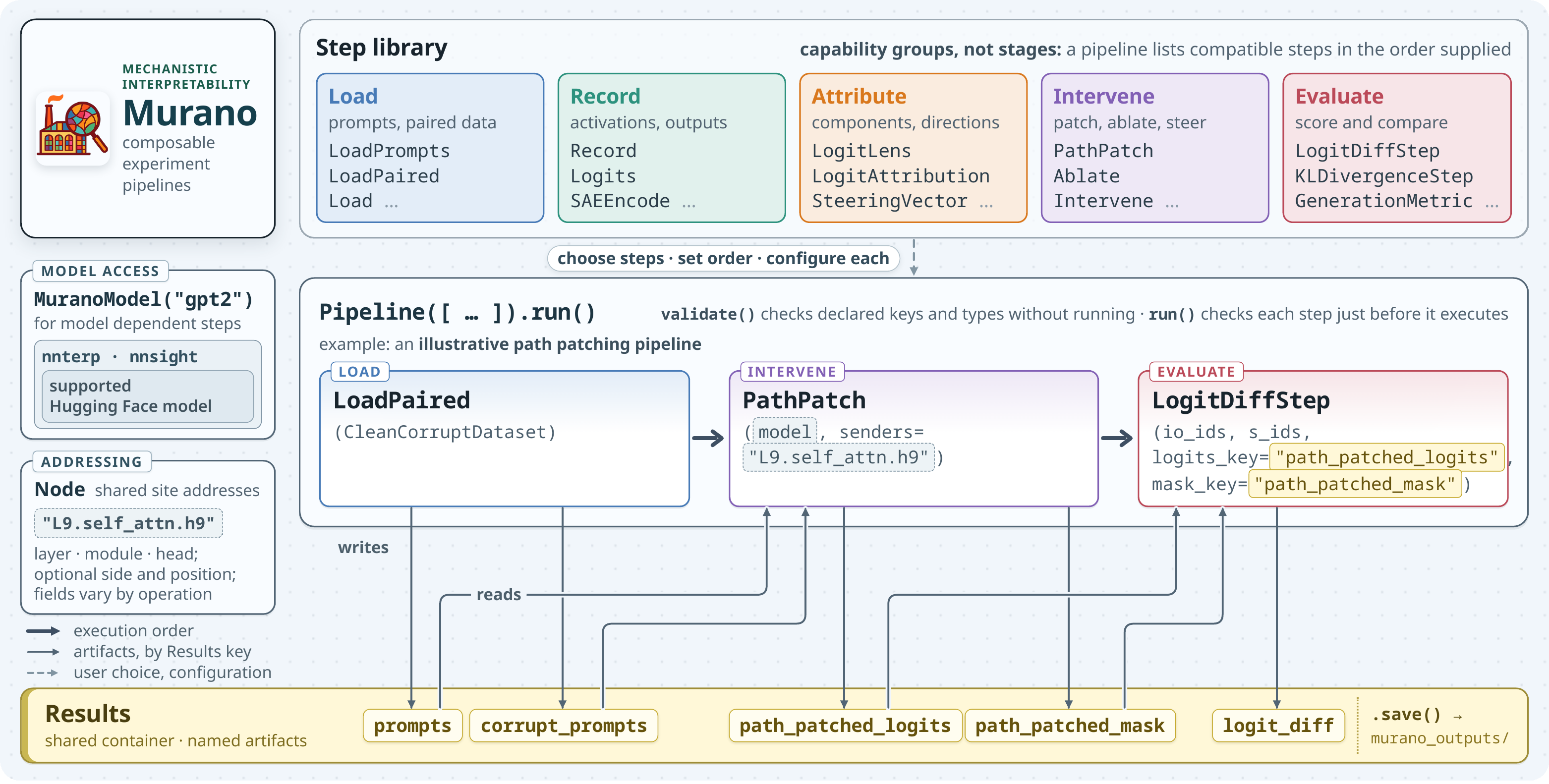}
  \caption{Murano pipeline architecture and an illustrative path patching workflow. A \texttt{Pipeline} executes compatible \texttt{Step} instances in the order supplied, and the steps exchange named artifacts through a shared \texttt{Results} container. Steps that depend on a model use \texttt{MuranoModel}, and supported operations accept canonical \texttt{Node} addresses. The five groups organize available capabilities rather than mandatory stages.}
  \label{fig:arch}
\end{figure*}

Many mechanistic interpretability studies combine three core operations: recording model activations, using attribution methods to localize components associated with a behavior, and intervening on those components to measure how a specified manipulation changes the behavior \citep{DBLP:journals/corr/abs-2407-02646, geiger2021causal}. The recording stage captures model activations without changing them. Attribution localizes heads, layers, directions, or sparse autoencoder features associated with a behavior. Without an intervention, this evidence is correlational. Interventions such as patching, steering, and ablation manipulate the attributed components or representations and measure the resulting behavioral changes. 

Existing libraries provide complementary capabilities: \texttt{TransformerLens}~\citep{nanda2022transformerlens} and \texttt{nnsight}~\citep{fiottokaufman2024nnsightndifdemocratizingaccess} expose model internals, \texttt{sae-lens}~\citep{bloom2024saetrainingcodebase} provides sparse autoencoders~\citep{DBLP:conf/iclr/HubenCRES24}, and \texttt{nnterp}~\citep{DBLP:journals/corr/abs-2511-14465} standardizes access to transformer modules across supported model families. Murano's design for composing configured operations is inspired by \texttt{Keras}~\citep{chollet2015keras} and \texttt{scikit-learn}~\citep{scikit}, adapted here to mutable experiment artifacts and interventions on model internals. Researchers who combine several such libraries may need to map one library's names for model components to another's, convert outputs into formats the next library can use, put dependent operations in the correct order, and connect intervention and evaluation code across scripts.

Murano is a framework for orchestrating mechanistic interpretability experiments through a shared step contract, result store, and model addressing scheme. For Hugging Face transformer architectures supported by \texttt{nnterp} and the relevant Murano operation, a pipeline is an ordered sequence of steps that may implement one or more of five conceptual phases: load, record, attribute, intervene, and evaluate (Figure~\ref{fig:arch}). Steps exchange named result artifacts under declared input and output contracts. Operations that pass component identities between analyses represent them as canonical \texttt{Node} addresses. This reduces renaming and conversion specific to individual libraries without implying that every constructor accepts every \texttt{Node} form. Murano wraps implemented methods as reusable steps that researchers can configure and combine for different experiments. Researchers can extend the framework by implementing a new \texttt{Step} with declared inputs and outputs.

Murano is designed for researchers across disciplines conducting mechanistic interpretability studies of large language models. Through its step interface, researchers can combine activation recording, attribution, intervention, and evaluation in exploratory analyses, new experiments, and reproducible case studies. Documentation is available on the project website,\footnote{\url{https://ukplab.github.io/murano/}} and Murano can be installed from PyPI as \texttt{murano-interp}.\footnote{\url{https://pypi.org/project/murano-interp/}} Our contributions are: \textbf{(1)} an open source orchestration framework for composing mechanistic interpretability operations; \textbf{(2)} declared result keys and optional type expectations, together with a canonical \texttt{Node} representation for exchanging component identities; and \textbf{(3)} two reproductions that recover selected reported findings plus one SAE feature steering case study.

\section{Background and Related Work}

Mechanistic interpretability aims to explain a neural network's behavior in terms of its internal components and computations \citep{olah2022mechanistic, elhage2021mathematical}. Transformers use a residual stream to carry information between layers. Attention heads and MLP blocks read from this shared state and write their outputs back to it, allowing the flow of information between model components to be represented as a graph \citep{elhage2021mathematical}. Causal abstraction provides a formal description of activation and path patching as interventions on a model's internal states \citep{geiger2021causal, JMLR:v26:23-0058}.

Activation patching takes an activation produced by one input, inserts it at the corresponding location during a run on another input, and measures how the output changes \citep{vig2020investigating, meng2022locating}. Path patching focuses the intervention on information transmitted from a chosen sender to a chosen receiver \citep{goldowsky2023localizing, wang2023ioi}. The logit lens projects intermediate residual states into vocabulary space, while direct logit attribution projects component outputs to estimate their contributions to a selected logit \citep{nostalgebraist2020interpretingGptThe, elhage2021mathematical}. Sparse autoencoders represent activations using learned sparse features \citep{DBLP:conf/iclr/HubenCRES24}, and activation steering changes model activations along a chosen direction \citep{arditi2024refusal}. Murano makes these methods composable within a single experiment.

A growing set of libraries supports these operations, but most focus on a particular abstraction layer or subset of the workflow. \texttt{TransformerLens} \citep{nanda2022transformerlens} is widely used for mechanistic interpretability but requires support specific to each architecture, which bounds its model coverage. \texttt{nnsight} \citep{fiottokaufman2024nnsightndifdemocratizingaccess} and \texttt{pyvene} \citep{wu2024pyvene} provide intervention interfaces for supported PyTorch models, while \texttt{nnterp} \citep{DBLP:journals/corr/abs-2511-14465} standardizes access to selected transformer modules across model families. Other systems focus on adjacent tasks: \texttt{EasyEdit} and \texttt{EasyEdit2} support knowledge and behavior editing \citep{wang-etal-2024-easyedit, xu2025easyedit2}, while Neuronpedia \citep{lin2023neuronpedia} provides a hosted interface for feature exploration and intervention. Combining systems from these different categories can require users to align model addresses, data formats, and intervention code.

Murano currently uses \texttt{nnterp} and \texttt{nnsight} for model access and integrates \texttt{sae-lens} \citep{bloom2024saetrainingcodebase} and \texttt{scikit-learn} \citep{scikit} for selected operations. It coordinates these implementations through shared \texttt{Step} and \texttt{Results} contracts and uses canonical \texttt{Node} addresses when component identities pass between operations. Section~\ref{sec:system-overview} describes these contracts and their support boundaries.

\section{System Overview}
\label{sec:system-overview}
A Murano \texttt{Pipeline} passes a mutable \texttt{Results} container through an ordered list of steps. Figure~\ref{fig:arch} groups available steps into five conceptual phases, but an individual pipeline may omit or repeat phases. The following paragraphs describe the result contracts, model backend, and model addressing scheme.

\begin{figure*}[t]
  \centering
  \includegraphics[width=\textwidth]{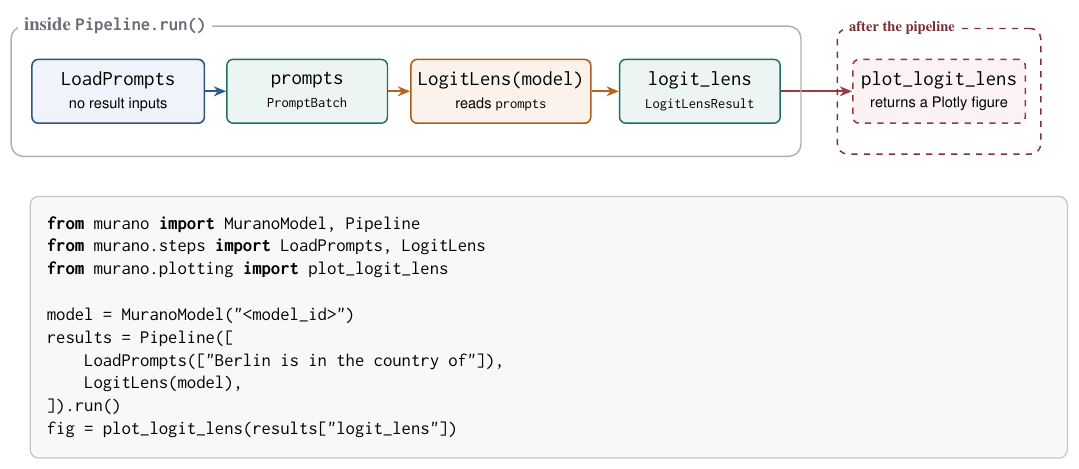}
  \caption{Artifact flow for a logit lens workflow. \texttt{LoadPrompts} stores a \texttt{PromptBatch} under \texttt{prompts}, and \texttt{LogitLens} reads it and stores a \texttt{LogitLensResult} under \texttt{logit\_lens}. Both operations run inside the pipeline. The plotting helper is called afterward with the stored result. The lower panel shows the corresponding executable code.}
  \label{fig:teaser}
\end{figure*}

\textbf{Results, steps, and pipelines.} The shared \texttt{Results} container maps string keys to arbitrary Python values; type expectations come from step declarations rather than the container itself. For example, \texttt{prompts} normally stores a \texttt{PromptBatch}, and \texttt{logit\_lens} stores a \texttt{LogitLensResult}.

The \texttt{Step} implementations included with Murano mutate and return the same \texttt{Results} container. A step declares the keys it reads and writes and may declare their expected types. \texttt{Pipeline.run()} checks required keys and declared input types immediately before executing each step. It does not check a produced value against the step's declared output type when that value is written.

Steps exchange run artifacts through declared \texttt{Results} keys, while step instances retain their configuration and model objects. If two \texttt{Step} instances in the same pipeline declare the same output key, Murano warns that the later value will replace the earlier one. For \texttt{Step} instances, \texttt{Pipeline.validate()} checks declared key availability and any available type declarations without executing the steps, but it assumes the pipeline starts with an empty \texttt{Results} container. \texttt{Pipeline.run()} can continue from an existing container, so an error in a later step may be found only after earlier steps have run.

\textbf{Model backend.} \texttt{MuranoModel} loads a Hugging Face model from an identifier or existing local path and wraps \texttt{nnterp}'s \texttt{StandardizedTransformer}, which is built on \texttt{nnsight}. Steps depend on the \texttt{ModelBackend} protocol, but its current interface retains \texttt{nnsight} proxy semantics and \texttt{MuranoModel} is its only implementation.

\textbf{Addressing model internals.} A \texttt{Node} stores a layer index and module name, with optional fields for an attention head, projection, and token position. It normalizes the aliases \texttt{residual}, \texttt{mlp\_out}, and \texttt{attn\_out}; other dotted module paths are stored as given and may remain architecture specific. In the canonical string form, \texttt{L5.self\_attn.h3.Q@p-1} denotes the Q projection of attention head 3 at token position $-1$ in layer 5; the projection field can likewise represent K, V, and O projections. Among current operations, only \texttt{PathPatch} uses the projection field, and only for Q/K/V receivers; other targeting operations reject it. Each operation validates the address forms and architectures it supports.

Artifacts that identify model components store canonical \texttt{Node} objects. \texttt{Ablate}, \texttt{Patch}, and \texttt{PathPatch} accept \texttt{Node} addresses and shorthand forms. \texttt{Record} takes separate layer, module, and position arguments, while \texttt{Intervene} takes separate layer and module arguments. The supported address fields therefore depend on the operation.

\textbf{Intervention.} Murano represents activation interventions with a callback \texttt{Callable[[Tensor, Node], Tensor]}. The backend applies the callback at selected module hooks during an intervened forward pass or throughout autoregressive generation. \texttt{Ablate} replaces selected activations with zeros, an estimated or supplied mean, or activations resampled from another example or source run. \texttt{Patch} configures resampling across runs: it runs one prompt batch while inserting aligned activations from another batch and writes the resulting logits. \texttt{Intervene} can project out a supplied direction or add a normalized direction with a chosen scale during generation. Downstream metric steps can then compare the resulting logits or generations.

\textbf{The step library.} The following classes implement the \texttt{Step} contract. \texttt{Record} captures activations from selected layers and modules, optionally retaining all token positions or separating attention heads. \texttt{RecordAttention} captures the full attention weight tensor for every head at each selected layer. \texttt{LogitLens} projects each selected layer's residual output through the model's final normalization and unembedding.

\texttt{SteeringVector} estimates a direction from the difference between the mean activations of two classes. \texttt{Intervene} can add the normalized direction at selected layers and modules during generation. \texttt{AblateAttention} replaces selected heads' attention weights with zeros, the batch mean pattern, or aligned weights from a source run and writes the resulting logits. For SAEs, \texttt{SAEEncode} records feature activations, helper functions rank candidate features, \texttt{SAETopActivations} and \texttt{SAEFeatureLabel} describe specified features from their activating contexts and promoted tokens, and \texttt{sae\_steer} constructs an \texttt{Intervene} step from one decoder direction. \texttt{Probe} evaluates logistic regression with cross validation by default and accepts another compatible \texttt{scikit-learn} estimator.

\texttt{LogitAttribution} estimates contributions from individual attention heads and MLPs after freezing the scale of the final normalization and reports the resulting completeness error. \texttt{PathPatch} freezes attention head outputs at their base values, inserts activations from a source run at specified sender heads or MLPs, captures the resulting activation at a \texttt{resid\_post} or head Q/K/V receiver, and writes the resulting logits. By default, MLPs can mediate the perturbation while the receiver activation is captured; \texttt{freeze\_mlps=True} freezes them during that capture to isolate the direct residual path to the receiver.

Steps exchange selected targets as canonical \texttt{Node} objects. \texttt{SelectComponents} stores these objects in a \texttt{ComponentSelection}. \texttt{Ablate} and \texttt{Patch} can read this artifact through \texttt{targets\_key}, and \texttt{PathPatch} can read its senders through \texttt{senders\_key}. Other operations may still require targets in their constructors.

\textbf{Evaluation, persistence, and optional dependencies.} \texttt{GenerationMetric} applies scorers supplied by users to paired baseline and modified generations. Separate steps compute logit difference, KL divergence, and recovered fractions from stored forward pass results. \texttt{Save} and \texttt{Results.save()} persist artifact types registered in \texttt{murano.io}. Values without a registered serializer are skipped; Murano warns for skipped nonprimitive values unless the value has already been recorded as metadata. Probing, dataset loading, plotting, SAE support, and notebook tools are optional extras.

\section{Usage}

Users construct a \texttt{Pipeline} by listing configured steps in execution order and then call \texttt{run()}. Figure~\ref{fig:teaser} shows a complete logit lens~\citep{nostalgebraist2020interpretingGptThe} pipeline for a supported Hugging Face model. \texttt{LoadPrompts} stores a \texttt{PromptBatch}; \texttt{LogitLens} projects each selected layer's residual output through the model's final normalization and unembedding and writes a \texttt{LogitLensResult}; \texttt{plot\_logit\_lens} visualizes that artifact.

Outside the pipeline API, \texttt{MuranoModel} provides convenience methods: \texttt{model.record(...)} performs a recording operation, while \texttt{model.generate(..., ablate=direction)} runs baseline and intervened generation; these methods do not define additional workflow phases. Named result keys make dependencies explicit. When a workflow does not depend on existing results, users can call \texttt{Pipeline.validate()} to check the declared key and type dependencies before running it.

We compare frameworks across the seven operation categories shown in Table~\ref{tab:compare} in Appendix~\ref{app:framework-comparison}. Recording captures model activations. Attribution estimates contributions from individual heads or MLPs, while activation patching replaces a target activation with one from a source run and measures a downstream change. Attention operations record or edit the weight matrix for each selected head. Steering adds selected directions, including SAE decoder directions, to the residual stream. Linear probes estimate whether labels are decodable from recorded activations. Among the libraries we compare against, Murano is the only one whose documented public interface directly implements all seven listed operation categories: recording, attribution, patching, attention analysis, steering, SAE analysis, and probing.

The same result and addressing contracts also support workflows with multiple steps. Feature steering, for example, loads a pretrained SAE through \texttt{sae-lens}, encodes activations with \texttt{SAEEncode}, ranks candidate features by activation on selected tokens with the \texttt{top\_sae\_features\_for\_tokens} helper, and steers generation with \texttt{sae\_steer}.

\section{Evaluation and Case Studies}
We use two published studies and one SAE feature steering case study to evaluate whether Murano can express the selected workflows. We report results from the Murano notebooks and count the logical statements in the corresponding core snippets, excluding imports, comments, printing, and shared setup. These counts describe the shown source only; they do not measure usability or reproducibility. The case studies do not evaluate user completion time, error rates, runtime, memory use, model coverage, the benefit of the \texttt{Node} representation by itself, or the effort required to add a new method.

\subsection{Indirect Object Identification}
\citet{wang2023ioi} identify attention heads that contribute to indirect object identification (IOI) in GPT-2 small~\citep{radford2019gpt2}, using sentences such as ``When Mary and John went to the store, John gave a drink to~\_\_.'' To test the reported heads, we replace each head's output at every prompt's final real token with the corresponding output from a corrupted prompt. The score is the batch mean of the correct indirect object logit minus the distractor subject logit. We report the relative change from the clean score as a percentage: $100 \times (s_{\mathrm{patched}} - s_{\mathrm{clean}}) / s_{\mathrm{clean}}$.

The three reported name mover heads, L9H9, L9H6, and L10H0, have the three strongest negative effects, while the two reported negative mover heads, L10H7 and L11H10, have the two strongest positive effects. For the three name mover heads, mean attention from the final token to the indirect object is 0.812, 0.742, and 0.460. In a separate output projection test, the tested name tokens appear among the five highest scoring tokens in 96\% to 100\% of cases. These measurements characterize the heads' attention patterns and copying tendency; they do not show that a head copied the answer in the original forward pass.

Query patches through the reported S-inhibition heads lower the logit difference for L9H9 and L10H0 by 28.9\% and 22.8\% of the clean to corrupt gap, but raise it for L9H6 by 4.9\%. Value patches produce no measurable change at the displayed precision. The S-inhibition result therefore holds for two of the three tested name mover heads. Mean ablation of the three name mover heads changes the logit difference from 3.484 to 3.520, so this result does not show that these heads are necessary for the answer.

For the attention head sweep, \texttt{PathPatch} captures the clean and corrupt head outputs, freezes all attention head outputs at their clean values, inserts the selected corrupt output, and runs the patched model. \texttt{LogitDiffStep} then computes the score. MLP outputs are not frozen in the reported sweep, so they may carry part of the resulting change. The corresponding core snippets are shown in Appendix Figure~\ref{lst:ioicmp}.

\subsection{Truth Direction}

The second study examines the truth directions described by \citet{marks2024geometrytruthemergentlinear}. The original study estimates directions that distinguish residual activations for true and false statements in LLaMA-2 models~\citep{touvron2023llama2openfoundation}. It adds or subtracts these directions and measures the change in the model's probability difference between \texttt{TRUE} and \texttt{FALSE}. Training on one group of statement types and testing on others measures how well the directions transfer.

\begin{figure*}[!t]
  \centering
  \includegraphics[width=\textwidth]{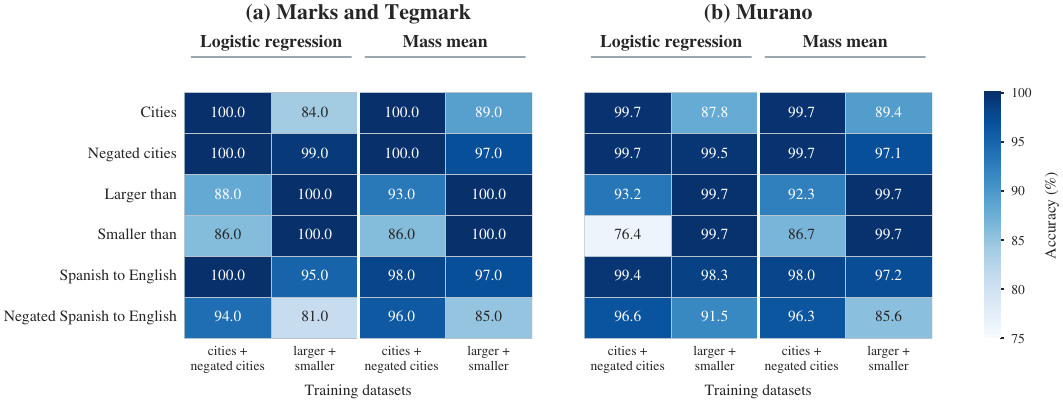}
  \caption{Probe accuracy in the truth direction reproduction. The left panel reports values from \citet{marks2024geometrytruthemergentlinear}, and the right panel reports Murano results. Logistic regression and mass mean probes are trained on the two data compositions shown and evaluated on six datasets. Both panels use the same scale from 75 to 100 percent. Cell values are rounded to one decimal place; the largest absolute difference computed from the unrounded results is 10.53 percentage points.}
  \label{fig:truth-gen}
\end{figure*}

We reproduce selected LLaMA-2-13B results with a Murano workflow. \texttt{MuranoDataset} stores separate groups of true and false statements. \texttt{Record} captures their layer 14 residual activations at each statement's final token, and \texttt{SteeringVector(normalize=False)} computes the true group mean minus the false group mean. The notebook defines a callback that adds or subtracts this vector at the two positions around the final period in layers 8 through 14, and \texttt{forward\_logits} runs the modified forward pass and returns the logits.

We reproduce the mass mean intervention trained on \texttt{cities+neg\_cities}. The normalized indirect effect measures the fraction of the baseline gap between the \texttt{TRUE} and \texttt{FALSE} probability scores covered by the intervention. Murano obtains 0.88 when adding the direction to false statements and 0.98 when subtracting it from true statements, compared with 0.85 and 0.97 in \citet{marks2024geometrytruthemergentlinear}. The largest difference is 0.03. Table~\ref{tab:reproduction-summary} summarizes these values together with selected IOI findings. Separately, Figure~\ref{fig:truth-gen} compares logistic regression and mass mean probe accuracy on six datasets and two training compositions; the largest difference across the 24 corresponding pairs of values is 10.53 percentage points.

\begin{table}[t]
  \centering
  \small
  \setlength{\tabcolsep}{3.5pt}
  \renewcommand{\arraystretch}{1.12}
  \begin{tabular}{@{}>{\raggedright\arraybackslash}p{0.25\columnwidth}>{\raggedright\arraybackslash}p{0.27\columnwidth}>{\raggedright\arraybackslash}p{0.39\columnwidth}@{}}
    \hline
    \textbf{Finding} & \textbf{Original} & \textbf{Murano} \\
    \hline
    \multicolumn{3}{@{}l}{\textbf{IOI reproduction}} \\
    Name mover heads & L9H9, L10H0, L9H6 & Three strongest: L9H9, L9H6, L10H0 \\
    Negative mover heads & L10H7, L11H10 & Two strongest: L10H7, L11H10 \\
    Name tokens in top 5 & 100\% & 96--100\% \\
    S-inhibition query patch & Effect reported & Lowers logit difference for L9H9 and L10H0; raises it for L9H6 \\
    S-inhibition value patch & No effect & No measurable change \\
    \hline
    \multicolumn{3}{@{}l}{\textbf{Truth direction: normalized indirect effect}} \\
    Add to false statements & 0.85 & 0.88 \\
    Subtract from true statements & 0.97 & 0.98 \\
    \hline
  \end{tabular}
  \caption{Selected IOI results from \citet{wang2023ioi} and truth direction results from \citet{marks2024geometrytruthemergentlinear}, alongside the Murano results. Truth direction setup: LLaMA-2-13B, \texttt{cities+neg\_cities} training, and \texttt{sp\_en\_trans} evaluation.}
  \label{tab:reproduction-summary}
\end{table}

Appendix Figure~\ref{lst:truthcmp} compares the core code for one additive mass mean pass: 10 logical statements with Murano and 14 with nnsight. \texttt{SteeringVector(normalize=False)} computes the raw difference between the group means, the callback states where to add it, and \texttt{forward\_logits} runs the intervention.

\subsection{Sparse Autoencoder Feature Steering}

We next use Murano to add the vector associated with one Gemma Scope feature~\citep{lieberum2024gemmascope} to the internal state of Gemma 2 2B Instruct~\citep{gemma2024} during generation. This case study is inspired by \citet{templeton2024scaling}, but it is not a reproduction because it uses a different model, SAE, concept, and intervention. Murano normalizes the selected feature vector and adds it with the chosen strength at layer 20 while each token is generated, whereas \citet{templeton2024scaling} clamp the feature activation.

We encode six sentences containing \texttt{California} with a Gemma Scope SAE and rank eight candidate features by their mean activation on the \texttt{California} tokens. Feature 1466 is the first candidate for which the model's output layer places three variants of \texttt{California} among the three tokens with the highest scores, so we use it for the intervention. This selection does not show that the feature responds only to California.

We evaluate four unrelated prompts at strengths 150, 420, and 2000. At strength 420, two displayed generations contain broken or repeated word fragments. At strength 2000, all four collapse into repetitions of \texttt{California} or \texttt{Sacramento}. These examples show that a large fixed addition can overwhelm generation; they do not show coherent control of topic.

Under our counting protocol, the shown core SAE workflow contains 13 logical statements with Murano and 24 in the direct \texttt{sae-lens} snippet. \texttt{SAEEncode} captures and encodes residual states, \texttt{top\_sae\_features\_for\_tokens} ranks candidates, \texttt{SAEFeatureLabel} lists their promoted output tokens, and \texttt{sae\_steer} constructs the additive intervention. This comparison does not cover feature clamping. The core snippets are shown in Appendix Figure~\ref{lst:saecmp}.

\enlargethispage{\baselineskip}
\section{Conclusion}
Murano coordinates selected mechanistic interpretability operations as ordered steps that exchange named results and use consistent addresses for model components. The IOI head sweep recovers the three reported name mover heads and two reported negative mover heads, although the S-inhibition query result holds for only two of the three tested name mover heads. In the truth direction case study, the two reproduced mass mean intervention values differ from the original values by at most 0.03, while the probe accuracy grid differs by at most 10.53 percentage points. The SAE case study shows that a large fixed feature addition can overwhelm generation. Murano encapsulates recurring implementation patterns as reusable operations. The case studies show how selected workflows can be expressed, and the truth direction notebook compares selected measurements with the original study. They do not directly evaluate readability, reproducibility, or extension effort.
 
\section*{Limitations}
Murano inherits the scientific assumptions of the methods it wraps and has additional system limitations: pipelines execute sequentially; the current backend depends on nnterp and nnsight; automatic GPU placement defaults to one GPU when CUDA is available; support for models and \texttt{Node} forms varies by operation and architecture; token positions depend on tokenization and padding; and analyses that retain many activations can require substantial accelerator memory. The current framework targets forward passes and bounded text generation with one model. Adapting it to language agents with tool use, persistent memory, and state across multiple turns remains future work. The case studies use fixed prompt sets, data splits, corruption schemes, and intervention settings. We do not systematically test sensitivity to alternative samples or hyperparameters. The framework comparison records documented interface coverage rather than implementation quality, usability, or performance.

\section*{Ethics Statement}
Murano is intended for research on understanding and controlling language model behavior. The same intervention capabilities could be misused to elicit harmful outputs or weaken safeguards. Users should follow applicable institutional policies, model licenses, and responsible disclosure practices.

\vfill
\newpage

\section*{Acknowledgement} 

This work was funded by the LOEWE Distinguished Chair ``Ubiquitous Knowledge Processing'', LOEWE initiative, Hesse, Germany (Grant Number: LOEWE/4a//519/05/00.002(0002)/81), as well as by the German Federal Ministry of Education and Research and the Hessian Ministry of Higher Education, Research, Science and the Arts within their joint support of the National Research Center for Applied Cybersecurity ATHENE, and by the Deutsche Forschungsgemeinschaft (DFG, German Research Foundation) under Germany's Excellence Strategy, EXC-3057. Federico Tiblias is supported by the Konrad Zuse School of Excellence in Learning and Intelligent Systems (ELIZA) through the DAAD programme Konrad Zuse Schools of Excellence in Artificial Intelligence, sponsored by the Federal Ministry of Education and Research.

\bibliography{custom}

\clearpage
\appendix

\setcounter{dbltopnumber}{2}
\renewcommand{\dbltopfraction}{1}
\renewcommand{\dblfloatpagefraction}{0.95}
\renewcommand{\textfraction}{0}
\makeatletter
\setlength{\@dblfptop}{0pt}
\setlength{\@dblfpsep}{4pt}
\setlength{\@dblfpbot}{0pt plus 1fil}
\makeatother

\begin{table*}[!t]

\refstepcounter{section}
\label{app:framework-comparison}
\noindent
{\large\bfseries \thesection\hspace{0.7em}Framework Capability Comparison}
\par\vspace{0.35em}

  \centering
  \resizebox{\textwidth}{!}{%
  \begin{tabular}{lccccccc}
    \hline
    \textbf{Framework} & \textbf{Record} & \textbf{Attribution} & \textbf{Patching} & \textbf{Attention} & \textbf{Steering} & \textbf{SAE} & \textbf{Probing} \\
    \hline
    \multicolumn{8}{l}{\textit{User libraries}} \\
    \textbf{Murano} & \cmark & \cmark & \cmark & \cmark & \cmark & \cmark & \cmark \\
    TransformerLens~\citep{nanda2022transformerlens} & \cmark & \cmark & \cmark & \cmark & \pmark & \pmark & \pmark \\
    nnsight~\citep{fiottokaufman2024nnsightndifdemocratizingaccess} & \cmark & \pmark & \cmark & \pmark & \pmark & \pmark & \pmark \\
    pyvene~\citep{wu2024pyvene} & \cmark & \xmark & \cmark & \pmark & \cmark & \cmark & \pmark \\
    \texttt{sae-lens}~\citep{bloom2024saetrainingcodebase} & \cmark & \pmark & \pmark & \pmark & \pmark & \cmark & \xmark \\
    EasyEdit2~\citep{xu2025easyedit2} & \pmark & \xmark & \xmark & \xmark & \cmark & \cmark & \xmark \\
    \multicolumn{8}{l}{\textit{Model access backend}} \\
    nnterp~\citep{DBLP:journals/corr/abs-2511-14465} & \cmark & \cmark & \cmark & \pmark & \cmark & \pmark & \pmark \\
    \multicolumn{8}{l}{\textit{Hosted platform}} \\
    Neuronpedia~\citep{lin2023neuronpedia} & \cmark & \pmark & \xmark & \xmark & \cmark & \cmark & \cmark \\
    \hline
  \end{tabular}}
  \captionof{table}{Capability comparison based on the cited publications and public documentation, checked in August 2026. For libraries and the backend, \cmark\ denotes a directly documented public interface, \pmark\ denotes an operation that requires user supplied hooks or code or a documented companion package, and \xmark\ denotes that no support was identified in the cited sources. For SAE use in the TransformerLens and nnsight rows, the companion package is \texttt{sae-lens}. For the hosted Neuronpedia row, \cmark\ denotes a documented platform feature. The groups distinguish user libraries from a model access backend and a hosted platform.}
  \label{tab:compare}

\end{table*}

\refstepcounter{section}

\begin{figure*}[!t]

\noindent
{\large\bfseries \thesection\hspace{0.7em}Evaluation Implementation Details}
\par\vspace{0.45em}

\centering

\begin{minipage}[t]{0.43\textwidth}
\centering
{\footnotesize\textbf{Murano}}\par\vspace{1pt}

\setcounter{clnum}{0}
\begin{lstlisting}[style=appendixcode]
(*@\cn@*)def logit_diff(res, logits_key, mask_key):
(*@\cn@*)    step = LogitDiffStep(correct=io_ids, incorrect=s_ids,
                         logits_key=logits_key, mask_key=mask_key)
(*@\cn@*)    return step(res)[keys.LOGIT_DIFF].value

(*@\cn@*)model = MuranoModel("gpt2", dtype=torch.float32,
                    enable_attention_probs=True)
(*@\cn@*)ds = CleanCorruptDataset(clean=clean, corrupt=corrupt,
                         correct=io_ids, incorrect=s_ids)
(*@\cn@*)base = LoadPaired(ds)(Results())
(*@\cn@*)final_positions = (
            base[keys.ATTENTION_MASK].sum(dim=1) - 1)
(*@\cn@*)clean_ld = logit_diff(Logits(model)(base),
                      keys.FINAL_LOGITS, keys.ATTENTION_MASK)
(*@\cn@*)effect = torch.zeros(model.n_layers, model.n_heads)

(*@\cn@*)for layer in range(model.n_layers):
(*@\cn@*)    for head in range(model.n_heads):
(*@\cn@*)        path_patch = PathPatch(
            model, Node(layer, SELF_ATTN, head=head),
            positions=final_positions)
(*@\cn@*)        out = path_patch(base)
(*@\cn@*)        ld = logit_diff(out, keys.PATH_PATCHED_LOGITS,
                        keys.PATH_PATCHED_MASK)
(*@\cn@*)        effect[layer, head] = 100 * (ld - clean_ld) / clean_ld

(*@\cn@*)top = effect.flatten().argsort()[:3]
(*@\cn@*)movers = [(int(i) // model.n_heads,
           int(i) % model.n_heads) for i in top]
\end{lstlisting}
\end{minipage}
\hfill%
\begin{minipage}[t]{0.55\textwidth}
\centering
{\footnotesize\textbf{nnsight}}\par\vspace{1pt}

\setcounter{clnum}{0}
\begin{lstlisting}[style=appendixcode]
(*@\cn@*)model = LanguageModel("openai-community/gpt2",
                      device_map="auto", dispatch=True)
(*@\cn@*)L = model.config.n_layer
(*@\cn@*)H = model.config.n_head
(*@\cn@*)D = model.config.n_embd // H
(*@\cn@*)B = len(clean)
(*@\cn@*)tokens = model.tokenizer(
            clean, padding=True, return_tensors="pt")
(*@\cn@*)final_positions = (
            tokens["attention_mask"].sum(dim=1) - 1)
(*@\cn@*)io = torch.tensor(io_ids)
(*@\cn@*)s = torch.tensor(s_ids)
(*@\cn@*)rows = torch.arange(B)

(*@\cn@*)def logit_diff(logits: torch.Tensor) -> torch.Tensor:
(*@\cn@*)    last = logits.detach().float().cpu()[
        rows, final_positions]
(*@\cn@*)    return (last[rows, io] - last[rows, s]).mean()

(*@\cn@*)def cache(prompts):
(*@\cn@*)    with model.trace(prompts):
(*@\cn@*)        heads = torch.stack([
            model.transformer.h[i].attn.c_proj.input
            for i in range(L)]).save()
(*@\cn@*)        logits = model.lm_head.output.save()
(*@\cn@*)    return heads, logits

(*@\cn@*)clean_heads, clean_logits = cache(clean)
(*@\cn@*)corrupt_heads, _ = cache(corrupt)
(*@\cn@*)clean_ld = logit_diff(clean_logits)
(*@\cn@*)effect = torch.zeros(L, H)

(*@\cn@*)for layer in range(L):
(*@\cn@*)    for head in range(H):
(*@\cn@*)        sender = clean_heads[layer].clone().reshape(
            B, -1, H, D)
(*@\cn@*)        corrupt_sender = corrupt_heads[layer].reshape(
            B, -1, H, D)
(*@\cn@*)        sender[rows, final_positions, head] = (
            corrupt_sender[rows, final_positions, head])
(*@\cn@*)        sender = sender.reshape(B, -1, H * D)

(*@\cn@*)        with model.trace(clean):
(*@\cn@*)            for i in range(L):
(*@\cn@*)                model.transformer.h[i].attn.c_proj.input = (
                    sender if i == layer else clean_heads[i])
(*@\cn@*)            patched = model.lm_head.output.save()

(*@\cn@*)        effect[layer, head] = (
            100 * (logit_diff(patched) - clean_ld) / clean_ld)

(*@\cn@*)top = effect.flatten().argsort()[:3]
(*@\cn@*)movers = [(int(i) // H, int(i) % H) for i in top]
\end{lstlisting}
\end{minipage}

\vspace{-3pt}
\caption{Core logic for the attention head sweep with Murano (17 logical statements, left) and a direct nnsight implementation (35 logical statements, right). Shared imports, prompts, labels, configuration, comments, and printing are omitted. Both snippets replace each selected head's output at the final real token, keep all other attention head outputs at their clean values, and use the same logit difference metric.}
\label{lst:ioicmp}

\end{figure*}

\begin{figure*}[!t]
\centering

\begin{minipage}[t]{0.39\textwidth}
\centering
{\footnotesize\textbf{Murano}}\par\vspace{1pt}

\setcounter{clnum}{0}
\begin{lstlisting}[style=appendixcode]
(*@\cn@*)model = MuranoModel(MODEL_ID)
(*@\cn@*)encode = SAEEncode(model, release=SAE_RELEASE, sae_id=SAE_ID)
(*@\cn@*)results = Pipeline([LoadPrompts(PROBES), encode]).run()
(*@\cn@*)record = results["sae_record"]

(*@\cn@*)cands = list(top_sae_features_for_tokens(
    record, model, {CONCEPT}, n=8))
(*@\cn@*)results = Pipeline([SAEFeatureLabel(
    model, feat_ids=cands, k_tokens=3)]).run(results)
(*@\cn@*)labels = results["feature_labels"]

(*@\cn@*)def matches(fid):
(*@\cn@*)    return any(CONCEPT.lower() in token.lower()
               for token in labels.tokens[fid])

(*@\cn@*)feat = next((fid for fid in cands if matches(fid)), cands[0])
(*@\cn@*)steer = sae_steer(model, encode.sae_model, feat, alpha=ALPHA)
(*@\cn@*)results = Pipeline([LoadPrompts(PROMPTS), steer]).run()
(*@\cn@*)gens = results["intervene"].modified_generations
\end{lstlisting}
\end{minipage}
\hfill%
\begin{minipage}[t]{0.59\textwidth}
\centering
{\footnotesize\textbf{Direct sae-lens}}\par\vspace{1pt}

\setcounter{clnum}{0}
\begin{lstlisting}[style=appendixcode]
(*@\cn@*)tok = AutoTokenizer.from_pretrained(MODEL_ID)
(*@\cn@*)model = AutoModelForCausalLM.from_pretrained(
    MODEL_ID, torch_dtype=torch.bfloat16, device_map="auto")
(*@\cn@*)sae = SAE.from_pretrained(
    release=SAE_RELEASE, sae_id=SAE_ID, device="cuda")[0]

(*@\cn@*)enc = tok(PROBES, return_tensors="pt", padding=True).to(model.device)
(*@\cn@*)with torch.no_grad():
(*@\cn@*)    resid = model(**enc, output_hidden_states=True
                  ).hidden_states[LAYER + 1]
(*@\cn@*)    acts = sae.encode(resid.to(sae.W_enc.dtype))

(*@\cn@*)mask = torch.tensor([
    [CONCEPT.lower() in tok.decode([token]).lower() for token in row]
    for row in enc["input_ids"]], device=acts.device)
(*@\cn@*)scores = (acts * mask.unsqueeze(-1)).sum((0, 1))
(*@\cn@*)scores /= mask.sum().clamp(min=1)
(*@\cn@*)cands = scores.topk(8).indices.tolist()

(*@\cn@*)unembed = model.get_output_embeddings().weight
(*@\cn@*)norm = model.model.norm
(*@\cn@*)feat = cands[0]

(*@\cn@*)for fid in cands:
(*@\cn@*)    promoted = (unembed @ norm(
        sae.W_dec[fid].to(model.dtype))).topk(3).indices.tolist()
(*@\cn@*)    if any(CONCEPT.lower() in tok.decode([token]).lower()
           for token in promoted):
(*@\cn@*)        feat = fid
(*@\cn@*)        break

(*@\cn@*)direction = sae.W_dec[feat] / sae.W_dec[feat].norm()

(*@\cn@*)def steer_hook(module, inputs, output):
(*@\cn@*)    return (output[0] + ALPHA * direction,) + tuple(output[1:])

(*@\cn@*)handle = model.model.layers[LAYER].register_forward_hook(steer_hook)
(*@\cn@*)gens = [tok.decode(model.generate(
    **tok(prompt, return_tensors="pt").to(model.device),
    **GEN_KWARGS)[0]) for prompt in PROMPTS]
\end{lstlisting}
\end{minipage}

\vspace{-3pt}
\caption{Core logic for ranking candidate SAE features, inspecting promoted output tokens, and constructing an additive generation intervention with Murano (13 logical statements, left) and direct \texttt{sae-lens} code (24 logical statements, right). Murano's \texttt{SAEEncode} and \texttt{SAEFeatureLabel} steps, \texttt{top\_sae\_features\_for\_tokens} helper, and \texttt{sae\_steer} step factory encapsulate residual capture, candidate ranking, inspection of promoted output tokens, and construction of the hook used during generation. Shared imports, data, configuration, comments, and printing are omitted.}
\label{lst:saecmp}

\vspace{1pt}

\begin{minipage}[t]{0.47\textwidth}
\centering
{\footnotesize\textbf{Murano}}\par\vspace{1pt}

\setcounter{clnum}{0}
\begin{lstlisting}[style=appendixcode]
(*@\cn@*)direction = Pipeline([
    Load(MuranoDataset(positive_texts=t_c + t_n,
                       negative_texts=f_c + f_n)),
    Record(model, layers=[PROBE_LAYER], position="last"),
    SteeringVector(normalize=False),
]).run()["steering"].direction_per_layer[(PROBE_LAYER, "residual")]

(*@\cn@*)true_id = model.tokenizer.encode(" TRUE")[-1]
(*@\cn@*)false_id = model.tokenizer.encode(" FALSE")[-1]
(*@\cn@*)length = len(model.tokenizer.encode("This statement is:"))

(*@\cn@*)def intervene(act, node):
(*@\cn@*)    act[:, -length - 1:-length + 1] += direction
(*@\cn@*)    return act

(*@\cn@*)logits = model.forward_logits(toks, fn=intervene,
                              layers=INTERVENE_LAYERS)
(*@\cn@*)probs = logits[:, -1].softmax(-1)
(*@\cn@*)p_diff = (probs[:, true_id] - probs[:, false_id]).mean()
\end{lstlisting}
\end{minipage}
\hfill%
\begin{minipage}[t]{0.51\textwidth}
\centering
{\footnotesize\textbf{nnsight}}\par\vspace{1pt}

\setcounter{clnum}{0}
\begin{lstlisting}[style=appendixcode]
(*@\cn@*)pos, neg = acts[labels == 1], acts[labels == 0]
(*@\cn@*)direction = pos.mean(0) - neg.mean(0)
(*@\cn@*)direction = direction / direction.norm()
(*@\cn@*)direction = ((pos.mean(0) - neg.mean(0)) @ direction) * direction

(*@\cn@*)sites = [(layer, offset)
         for layer in range(8, 15)
         for offset in (-1, 0)]
(*@\cn@*)true_id = tok.encode(" TRUE")[-1]
(*@\cn@*)false_id = tok.encode(" FALSE")[-1]
(*@\cn@*)length = len(tok.encode("This statement is:"))

(*@\cn@*)with model.forward(remote=remote) as runner:
(*@\cn@*)    with runner.invoke(batch):
(*@\cn@*)        for layer, offset in sites:
(*@\cn@*)            model.model.layers[layer].output[0][
                :, -length + offset, :] += direction

(*@\cn@*)        probs = model.lm_head.output[:, -1].softmax(-1)
(*@\cn@*)        p_diff = (
            probs[:, true_id] - probs[:, false_id]
        ).save()
\end{lstlisting}
\end{minipage}

\vspace{-3pt}
\caption{Core logic for one additive mass mean intervention with Murano (10 logical statements, left) and nnsight (14 logical statements, right). Both snippets use a direction trained on the \texttt{cities+neg\_cities} groups. \texttt{SteeringVector(normalize=False)} computes the raw difference between the group means, the callback states where to add it, and \texttt{forward\_logits} runs the intervention. Shared imports, data aliases, configuration, comments, and printing are omitted.}
\label{lst:truthcmp}

\vspace{0pt}

\end{figure*}

\end{document}